\documentstyle[11pt,colap]{article}
\begin{document}
\title{Centering in Dynamic Semantics}

\author{ Daniel Hardt \\
Department of Computing Sciences \\
Villanova University \\Villanova, PA 19085\\
{\em hardt@vill.edu}}

\maketitle
\begin{abstract}
Centering theory posits a discourse center, a distinguished
discourse entity that is the topic of a discourse.  A simplified
version of this theory is developed in a Dynamic Semantics 
framework.  In the resulting system, the mechanism
of {\em center shift} allows a simple, elegant analysis of 
a variety of phenomena involving sloppy identity in ellipsis and ``paycheck
pronouns''.
\end{abstract}


\newcounter{sentencectr}
\newcounter{sentencesubctr}

\renewcommand{\thesentencectr}{(\smainform{sentencectr})}
\renewcommand{\thesentencesubctr}{\thesentencectr\ssubform{sentencesubctr}}

\newcommand{\smainform}{\arabic}
\newcommand{\ssubform}{\alph}
\newcommand{\ssubpunc}{.{}}

\newcommand{\beginsentences}{ %
\begin{list}{(\thesentencectr)}
   {\usecounter{sentencesubctr}
    \setlength{\labelwidth}{0.5 in}
    \addtolength{\leftmargin}{25 pt}
    \setlength{\parsep}{0 in}}} 
\def\endsentences{\end{list}}

\newcommand{\sitem}{\renewcommand{\thesentencesubctr}{(\smainform{sentencectr}}
                    \refstepcounter{sentencectr}
     \item[(\smainform{sentencectr})\hfill]}
\newcommand{\smainitem}{\renewcommand{\thesentencesubctr
                                    }{\thesentencectr\ssubform{sentencesubctr}}
                        \setcounter{sentencesubctr}{0}
                        \refstepcounter{sentencectr}
                        \refstepcounter{sentencesubctr}
     \item[\thesentencectr\hfill\ssubform{sentencesubctr}\ssubpunc]}
\newcommand{\ssubitem}{\refstepcounter{sentencesubctr}
     \item[\hfill\ssubform{sentencesubctr}\ssubpunc]}

\makeatletter            
\newcommand{\smainlabel}[1]{{
\renewcommand{\@currentlabel}{\thesentencectr}\label{#1}}}

\newcommand{\ssublabel}[1]{{
\renewcommand{\@currentlabel}{\ssubform{sentencesubctr}}\label{#1}}}
\makeatother

\makeatother

\newcommand{\ldb}{[\![}
\newcommand{\rdb}{]\!]}
\newcommand{\mo}[1]{\ldb {#1}\rdb}

\maketitle

\section{Introduction}  

Centering \cite{GJW:95} and Dynamic Semantics\footnote{The framework is due
to \cite{Kamp:80,Heim:82}; important subsequent papers include
\cite{Groenendijk&Stokhof:91,Chierchia:92}, and many others.}
both concern the sequential processing
of discourses, with particular emphasis on the resolution of pronouns.  
In Dynamic Semantics, the semantic structure of a discourse gives
rise to constraints on the resolution of anaphoric expressions.
Centering theory claims that a discourse always has a single topic,
or center.  Constraints on the resolution of anaphoric
expressions arise, in part, from the ways in which the center can
change in a discourse.  
There is an important difference in the way discourses are
viewed in Centering and in Dynamic Semantics.  In Dynamic Semantics,
a discourse is viewed as a monotonic increase in information, as
discourse referents are constantly added to the domain of 
discourse.  Centering draws attention to a particular {\em role} that
a discourse entity can hold; from time to time, the current center
will be {\em shifted} with a new center.  In this paper, I will
implement a simplified version of the centering theory in a dynamic
system, and I
will apply the resultant framework to a variety of phenomena involving
sloppy identity in ellipsis and ``paycheck
pronouns''.

Since Montague, a major goal of semantics has been to describe 
a compositional method for converting a syntactic representation of
a sentence into a logical representation of the sentence meaning,
and then to evaluate that representation with respect to a given
context.  A primary insight of dynamic semantics is that sentences
have a systematic relation to context in two ways: not only
are they evaluated with respect to the current context, but they
also systematically change that context.  
This insight has particular relevance for the apparent puzzle presented by
sloppy identity and related phenomena.
While anaphoric expressions are normally
thought to be identical in meaning to their antecedents, they receive
a different interpretation than their antecedents in these cases.
Given the dynamic perspective, the puzzle evaporates: the anaphoric
expression and its antecedent might represent exactly the same
meaning, since meaning is fundamentally a potential to be evaluated
with respect to some context.  What changes is the context, in the
discourse intervening between antecedent and anaphoric expression.

Consider the following example involving sloppy identity in VP ellipsis:

\beginsentences
\sitem Tom$_1$ loves his$_1$ cat.  John$_1$ does too. [loves his$_1$ cat]
\label{tomjohn0}
\endsentences

The sloppy reading results from a change in context, in which the value
of 1 becomes {\em John} rather than {\em Tom}.  This allows an extremely
simple account of the ``recovery mechanism'' involved in sloppy identity;
the elided VP is exactly identical to its antecedent. 
Several authors \cite{Gardent:91,Hardt:94} have suggested a dynamic
account along these lines, arguing 
that sloppy identity and related phenomena  reflect the {\em reassignment} 
of an index in the discourse context.\footnote{\cite{Kennedy:93} has
argued from a GB perspective for a similar approach to indexing,
in which sloppy identity involves exact identity, and a reassignment
of the controller index.}

Alternative approaches postulate complex recovery mechanisms for sloppy identity, such
as higher-order matching \cite{DSP:91} or the syntactic matching 
of parallel dependencies \cite{Fiengo&May:94}. Below, I will argue that
the dynamic account is more general and empirically adequate, as well
as being simpler than alternative accounts.  

The dynamic account raises the following problem: since
the index of the the initial ``controller'' is {\em reassigned}, it 
becomes inaccessible in subsequent discourse.  In the
above example, {\em Tom} would be rendered inaccessible, since its
position in the assignment function has be reassigned to {\em John}.
The notion of discourse center provides a solution to this problem.  The
index 0 will be reserved for the discourse center, and the discourse
center will always occupy another index as well as 0.  We will 
use the * to designate references to the discourse center.  Thus
the above example will be notated as follows:

\beginsentences
\sitem Tom$_{1*}$ loves his$_*$ cat.  John$_{2*}$ does too. [loves his$_*$ cat]
\label{tomjohn1}
\endsentences

In the first sentence, {\em Tom} is the value of index 1, and is also the
discourse center, i.e., the value of index 0.  The pronoun {\em his*}
is equivalent to his$_0$, and thus refers to the discourse center.  In
the second sentence, {\em John} becomes the value of index 2, and
also replaces {\em Tom} as the discourse center and thus {\em John}
becomes the value of index 0.  This {\em center shift} gives rise to 
the sloppy reading.  However, both {\em Tom} and {\em John} remain
accessible in subsequent discourse.  

The paper is organized as follows: In Section Two, I present a dynamic
framework based on the system described in \cite{Muskens:95}, with
extensions for  the discourse center, VP ellipsis, and paycheck pronouns.
Section Three concerns an ``expanded paradigm'' for sloppy identity;
it is shown that the proposed approach uniformly
accounts for a broad range of sloppy identity
phenomena, including some not previously examined in the literature.
Conclusions and plans for future work are given in Section Four.

\section{A Dynamic Framework}

The basic dynamic framework is the dynamic logic system of \cite{Muskens:95}.   
This framework has, for the sake of simplicity,  
restricted the study of anaphora to pronouns that
are extensionally identified with their antecedents\footnote{There
are several researchers who have extended dynamic frameworks to account for ellipsis
and related phenomena: \cite{Klein:84} is an early example.  \cite{Asher:93}
examines a variety of extensions to the DRT framework. 
\cite{Frances&vanEijk:93} explore similar issues of indexing and
ellipsis in a dynamic setting.  \cite{Gardent:91} also 
extends a dynamic semantics system for ellipsis and anaphora.}.  
I will extend Muskens' system to permit
anaphora involving VP's as well as NP's, and to allow antecedents
to be dynamic as well as ordinary (extensional) objects.

In Muskens' system, linearized DRT boxes are integrated with the type
logic \cite{Church:40} that underlies Montague Semantics. 
Linearized DRT boxes are simply a more concise way of writing
standard DRT boxes \cite{Kamp:80}.
Muskens shows that DRT boxes can be viewed as abbreviations for
expressions in ordinary type logic. 
Consider the following discourse:
the discourse:
{\em A$_1$ farmer walks. He$_1$ laughed.}

This is represented by the following linearized DRT box:

\begin{quotation}
[u$_1$ $\mid$ farmer(u$_1$), walk(u$_1$),laugh(u$_1$)]
\end{quotation}

This is an abbreviation for the following type logic formula:

\bigskip

\noindent$\lambda$ij(i[u$_1$]j$\wedge$farmer(u$_1$j)$\wedge$walks(u$_1$j)$\wedge$laughs(u$_1$j))


\bigskip

In the above formula, the variables $i$ and $j$ represent input and output 
states, and the variable u$_1$ (akin to a discourse marker) 
is a function from states to individuals.
In what follows, we use the DRT abbreviations without further comment.
The reader is referred to \cite{Muskens:95} for further examples
and the details of the system.

We now define a simple fragment of English, based on the one
given in \cite{Muskens:95}.

\begin{tabbing}
a$_n$ xxxx\= xxxx \= \kill
a$_n$ \> $\Rightarrow$ \> $\lambda$ P$_1$ P$_2$([u$_n$$\mid$];P$_1$(u$_n$);P$_2$(u$_n$)) \\
John$_n$ \> $\Rightarrow$ \> $\lambda$P([u$_n$ $\mid$ u$_n$ = John];P(u$_n$)) \\
he$_n$ \> $\Rightarrow$ \> $\lambda$P P($\delta$) where $\delta$=dr(ant(he$_n$)) \\
if \> $\Rightarrow$ \> $\lambda$pq [ $\mid$ p$\Rightarrow$q] \\
and \> $\Rightarrow$ \> ; \\ 
walk\> $\Rightarrow$ \> $\lambda$v [ $\mid$ walk(v)] \\
cat\>  $\Rightarrow$ \> $\lambda$v [ $\mid$ cat(v)] \\
love\> $\Rightarrow$ \>  $\lambda$Q $\lambda$v (Q($\lambda$u$^\prime$[ $\mid$ loves(v,u$^\prime$)])) \\ 
\end{tabbing}

Note that the translation for $he_n$ refers to $dr(ant(he_n))$.  
This is defined as the discourse representation of the antecedent of
$he_n$(see \cite[page 20]{Muskens:95}).  The translation
for $and$ is the {\em sequencing operator}, {\em ;}.  As described
in \cite{Muskens:95}, the sequencing of two boxes K,K$^\prime$ is 
an abbreviation for the following type logic expression:

\bigskip

\noindent $\mo{K_1;K_2}$ $\Rightarrow$ \\
\indent \{$<$i,j$>$ $\mid$ $\exists$k
($<$i,k$>$ $\epsilon$ $\mo{K_1}$ \&
$<$k,j$>$ $\epsilon$ $\mo{K_2}$)\}
	
\bigskip

Typically, two DRT boxes appearing in sequence can be {\em merged}
into a single box, consisting of the union of the discourse markers
in the two boxes and the union of the conditions.  This is described
in the {\em Merging Lemma} of \cite[page 8]{Muskens:95}.  In the
representations that follow, we will often merge boxes without
comment to simplify representations. However, the merge of two
boxes is not always possible -- if there is a reassignment of
an index, it will not be possible to perform the merge.  This
will arise in the cases of sloppy identity examined below.

The above fragment, following the Kamp/Heim accounts, considers only one 
type of anaphora, involving individuals.  We will extend the
fragment in the following ways:
\begin{itemize}
\item we will add the idea of a {\em discourse center} to the system
\item  we will allow
dynamic properties to be added to contexts, as antecedents for
VP ellipsis
\item we will allow dynamic individuals 
to be added to contexts, to account for ``paycheck pronouns''
\end{itemize}

\subsection{Discourse Center}

We define position 0 in the context as the {\em Discourse Center}.
At any given point in the discourse, the discourse entity designated
as the discourse center occupies position 0 as well as its other 
position.  We designate this with a *, as in the following example:

\beginsentences
\sitem A$_1$* farmer walks. He* laughed.
\endsentences

This is represented as follows:

\begin{tabbing}
 \= x u$_0$, u$_1$, P$_2$ $\mid$ \= \kill

[u$_0$,u$_1$ $\mid$ u$_0$ = u$_1$, farmer(u$_1$),
walk(u$_1$),laugh(u$_1$)]

\end{tabbing}

In this discourse, the entity introduced by {\em A$_1$* farmer} is the
discourse center, and thus occupies position 0 as well as position 1.

We must add additional rules for 
indefinite expressions and names, when they add an object to context that is the
discourse center.

\begin{tabbing}
a$_n$ xxxx\= xxxx \= \kill
a$_n$* \> $\Rightarrow$ \\
       \>  $\lambda$ P$_1$ P$_2$([u$_0$,u$_n$$\mid$ u$_0$ = u$_n$];
          P$_1$(u$_n$);P$_2$(u$_n$)) \\
John$_n$* \> $\Rightarrow$ \\
       \> $\lambda$P([u$_0$,u$_n$ $\mid$ u$_0$ = u$_n$,u$_n$ = John];P(u$_n$)) \\

\end{tabbing}

We will apply a very simplified version of centering theory, consisting
of the following constraints:

\begin{itemize}
\item Every discourse utterance (except the discourse initial utterance)
must have a center.
\item If any pronouns occur in an utterance, at least one pronoun must
refer to the center.
\end{itemize}

We define two types of {\em transitions} from one utterance to the next:
\begin{enumerate}
\item {\em Center Continuation}: the center remains the same
\item {\em Center Shift}: the center changes
\end{enumerate}

The actual centering theory involves an additional data structure,
the {\em forward-looking centers}, and defines four transition types,
with a preference ordering among them.  The reader is
referred to \cite{GJW:95} for a full account of this.  For our purposes, we will 
rely on the mechanism of {\em center shift} to implement the reassignment
that we argue is crucial to the dynamic account of sloppy identity.

\subsection{VP Ellipsis}

Next, we extend the system for VP ellipsis:
first, verbs are separated into a base form and an inflection 
(INFL).  This facilitates the treatment of VP ellipsis; the INFL category 
adds the new property to 
the context, just as the determiner ``a'' adds a new individual to the 
context. An alternative meaning for the INFL category is given for
VPE occurrences, where a property is accessed from the input context.
\begin{tabbing}
a$_n$ xxxx\= xxxx \= \kill
INFL$_n$ \> $\Rightarrow$ \> $\lambda$ P $\lambda$x
[P$_n$ $\mid$ P$_n$ = P] ; P(x) \\
INFL$_n$ \> $\Rightarrow$ \> dr(ant(INFL$_n$))
\end{tabbing}

The INFL category ranges over verbal inflections (PAST, PRES, etc.) 
and auxiliary verbs (do, should, etc.)\footnote{We
ignore the semantic contribution of INFL, apart from the above-described
interaction with the discourse context.}

Consider the following example of VP ellipsis:
\beginsentences
\smainitem Tom walks.  John does too.
\ssubitem Tom$_1$* PRES$_2$ walk.  John$_3$* does$_2$ too.
\endsentences

The two sentences receive the following interpretations:
\begin{tabbing}
 \= x u$_0$, u$_1$, P$_2$ $\mid$ \= \kill
 Tom$_1$* PRES$_2$ walk.  $\Rightarrow$ \\
 \> [u$_0$, u$_1$, P$_2$ $\mid$ u$_0$ = u$_1$, u$_1$ = Tom, \\ 
 \>  \> P$_2$ = $\lambda$ x[$\mid$ walk(x)], walk(u$_1$) ] \\

John$_3$* does$_2$ VPE$_2$ too.   $\Rightarrow$ \\
\> [u$_0$, u$_3$ $\mid$ u$_0$ = u$_3$, u$_3$ = John] ; P$_2$(u$_3$) \\
\end{tabbing}

Next, we join the two sentence interpretations with the sequencing operator,
and we apply the value of P$_2$ to u$_3$:

\begin{tabbing}
\= x u$_0$, u$_1$, P$_2$ $\mid$ \= \kill
 Tom$_1$* PRES$_2$ walk. John$_3$* does$_2$ VPE$_2$ too.   $\Rightarrow$ \\
\> [u$_0$, u$_1$, P$_2$ $\mid$ u$_0$ = u$_1$, u$_1$ = Tom, \\
\> \> P$_2$ = $\lambda$ x[$\mid$ walk(x)], walk(u$_1$) ] ; \\
\>  [u$_0$, u$_3$ $\mid$ u$_0$ = u$_3$, u$_3$ = John, walk(u$_3$)] \\
\end{tabbing}

Next, we will consider an example involving sloppy identity.
To do this, it will be necessary to add genitive
NP's, such as ``his cat'' to our system.

\begin{tabbing}
a$_n$ xxxx\= xxxx \= \kill
his (he$_n$'s$_m$) $\Rightarrow$ \\
 $\lambda$P$_1$P$_2$ ([u$_m$ $\mid$ of(u$_m$, u$_n$)]; P$_1$(u$_m$); P$_2$(u$_m$) ) \\
\end{tabbing}

We need two indices: $n$ is the index of $he$: this is an individual
defined in input context.  The index $m$ is the index of the object
possessed by $he_n$; this object is added to the output context.
(For clarity, we will often write $his_n cat_m$; but the ``official
usage'' is {\em he$_n$'s$_m$ cat}.)

Now, we examine a simple case of sloppy identity in VP ellipsis:

\beginsentences
\smainitem Tom loves his cat.  John does too.
\label{tomjohn}
\ssubitem Tom$_1$* PRES$_2$ love his* cat$_3$.  John$_4$* does$_2$ too.
\endsentences

\begin{tabbing}
[u$_0$, u$_1$, P$_2$, u$_3$ $\mid$ \= u$_0$ = u$_1$, u$_1$ = \= Tom, \kill

Tom$_1$* PRES$_2$ love his* cat$_3$ $\Rightarrow$ \\ 
 $[$ u$_0$, u$_1$, P$_2$, u$_3$ $\mid$ u$_0$ = u$_1$, u$_1$ = Tom, \\
 \> P$_2$ = $\lambda$x([u$_3$$\mid$ of(u$_3$, u$_0$), \\
  \> \> cat(u$_3$), love(x,u$_3$)]), \\
 \> of(u$_3$,u$_0$),cat(u$_3$), love(u$_1$,u$_3$)$]$ \\

John$_4$* does$_2$ (too) $\Rightarrow$ \\ 
 $[$ u$_0$, u$_4$ $\mid$ u$_4$ = u$_0$, u$_4$ = John$]$ ;
 P$_2$(u$_4$) \\
\end{tabbing}

Next, we join the two sentences together and apply the value
of P$_2$ to u$_4$:
\begin{tabbing}
[u$_0$, u$_1$, P$_2$, u$_3$ $\mid$ \= u$_0$ = u$_1$, u$_1$ = \= Tom, \kill

Tom$_1$* PRES$_2$ love his* cat$_3$ (and) \\
 John$_4$* does$_2$ (too) $\Rightarrow$ \\ 

$[$u$_0$, u$_1$, P$_2$, u$_3$ $\mid$ u$_0$ = u$_1$, u$_1$ = Tom, \\
\> P$_2$ = $\lambda$x[u$_3$$\mid$ of(u$_3$, u$_0$), \\
\> \> cat(u$_3$), love(x,u$_3$)],\\
\> of(u$_3$,u$_0$),cat(u$_3$), love(u$_1$,u$_3$)$]$ ;\\
 $[$u$_0$, u$_4$ $\mid$ u$_4$ = u$_0$, u$_4$ = John$]$ ; \\
$[$u$_3$ $\mid$ of(u$_3$, u$_0$), cat(u$_3$), love(u$_4$,u$_3$)$]$ \\

\end{tabbing}

The antecedent for the VPE is ``love his cat''.  This object (P$_2$) is 
introduced into the context by PRES$_2$.
P$_2$ represents the property of ``loving u$_0$'s cat'', where u$_0$ is the
discourse center defined in the input context.  In the first sentence, the 
center is TOM.  The second sentence {\em shifts} the center to JOHN.  It is
this change in context that gives rise to the sloppy reading.
Thus a sloppy reading is made possible when there is a {\em center shift}.

Finally, we allow the possibility that a property might be the
discourse center.  This means we must add an alternative rule
for INFL, so that it adds a property that is the discourse center:

\begin{tabbing}
a$_n$ xxxx\= xxxx \= \kill
INFL$_n$* \> $\Rightarrow$ \\
\> $\lambda$ P $\lambda$x
[P$_n$ $\mid$ P$_0$ = P$_n$, P$_n$ = P] ; P(x) \\

\end{tabbing}

\subsection{Paycheck Pronouns}

The phenomenon of ``paycheck pronouns'',\footnote{This term comes from Kartunnen's
example: {\em The man who gave his paycheck to his wife was wiser than the one who gave
it to his mistress.}  Various accounts of this phenomenon have been proposed, such
as \cite{Cooper:79,Engdahl:86,Jacobson:92,Gardent:91}. 
\cite{Heim:90} proposed extending the Sag/Williams account of VPE
to the case of paycheck pronouns.  Gardent makes a proposal similar to the
current account: a dynamic approach in which
paycheck pronouns and VPE are treated uniformly.}
is illustrated by the following example
\beginsentences
\sitem Smith spent his paycheck. Jones saved it.
\endsentences

The reading of interest is where the pronoun ``it'' refers to
Jones' paycheck, although its antecedent (``his paycheck'') 
refers to Smith's paycheck.
Our account for this parallels the account of sloppy identity in
VP ellipsis.  The antecedent ``his$_i$ paycheck'' introduces a {\em dynamic
individual}: a relation between contexts that introduces $i$'s paycheck
to the output context, where the value of $i$ is determined by the
input context.  The following rule makes it possible for NP's 
like ``his paycheck'' to add dynamic individuals to the context.

\begin{tabbing}
$\lambda$ P$_1$ P$_2$ \= [x$_m$ $\mid$ x$_m$ = $\lambda$P \= \kill
his (he$_n$'s$_m$) $\Rightarrow$ \\
 $\lambda$ P$_1$ P$_2$ [x$_m$ $\mid$ x$_m$ = $\lambda$P ([u$_m$ $\mid$ of(u$_m$,u$_n$)];\\
\> \>     P$_1$(u$_m$);P(u$_m$)); \\
\> x$_m$(P$_2$) \\

\end{tabbing}

We use variables of the form $u_i$ to denote ordinary {\em extensional} 
individuals; we use variables of the form $x_i$ to denote {\em dynamic}
individuals.  There are two distinct effects on the output context.
First, the dynamic individual x$_m$ is added to context: this object
adds an individual u$_m$ to a given context, such that u$_m$ is 
{\em of} u$_n$ in that context.  Second, x$_m$ is {\em applied} to the property
P$_2$.  This actually adds u$_m$ to the current context.

Finally, we need an alternative form for pronouns that
refer to dynamic individuals:

\begin{tabbing}
a$_n$ xxxx\= xxxx \= \kill
he$_n$ \> $\Rightarrow$ \> $\delta$ where $\delta$ = dr(ant(he$_n$)) \\
\end{tabbing}

The pronoun he$_n$ recovers x$_n$ from the current context.
The desired reading can now be derived as follows:
\bigskip
\beginsentences
\smainitem Smith spent his paycheck.
Jones saved it.
\ssubitem Smith$_1$* PAST$_2$ spend his* paycheck$_3$. Jones$_4$* PAST$_5$ save it$_3$.
\endsentences

We take the two sentences individually.  The first sentence introduces the
dynamic individual x$_3$, as follows\footnote{To simplify the representation,
we omit the values for VP variables P$_2$ and P$_5$, since they are
not relevant to the current example.}:

\begin{tabbing}
\= xx P$_2$ [x$_3$ $\mid$ \= x$_3$ = \= \kill

his* paycheck$_3$.  $\Rightarrow$\\
$\lambda$P$_2$ [x$_3$ $\mid$ x$_3$ = $\lambda$P([u$_3$ $\mid$ of(u$_3$,u$_0$), paycheck(u$_3$)]; \\
\>   \> \> P(u$_3$)) ]; \\
\>   \> x$_3$(P$_2$) \\

spend his* paycheck$_3$.  $\Rightarrow$\\
$\lambda$v [x$_3$ $\mid$ x$_3$ = $\lambda$P([u$_3$ $\mid$ of(u$_3$,u$_0$), paycheck(u$_3$)];\\
\> \> \> P(u$_3$)) ]; \\
\> \> x$_3$($\lambda$u$^\prime$[ $\mid$ spend(v,u$^\prime$)]) \\

spend his* paycheck$_3$.  $\Rightarrow$\\
$\lambda$v [x$_3$ $\mid$ x$_3$ = $\lambda$P([u$_3$ $\mid$ of(u$_3$,u$_0$), paycheck(u$_3$)];\\
\> \> \> P(u$_3$)) ]; \\
\>[u$_3$ $\mid$ of(u$_3$,u$_0$), paycheck(u$_3$)];[ $\mid$ spend(v,u$_3$)] \\

Smith $_1$* PAST$_2$ spend his* paycheck$_3$.  $\Rightarrow$\\
\> [u$_0$,u$_1$,P$_2$,x$_3$ $\mid$ u$_0$ = u$_1$,u$_1$ = Smith,\\
\> \> x$_3$ = $\lambda$P([u$_3$ $\mid$ of(u$_3$,u$_0$),paycheck(u$_3$)];\\
\> \> P(u$_3$))];  \\
\>[u$_3$ $\mid$ of(u$_3$,u$_0$), paycheck(u$_3$),spend(u$_1$,u$_3$)] \\

\end{tabbing}

We continue with the second sentence. 

\begin{tabbing}
xxx\= xxx\= \kill
save it$_3$  $\Rightarrow$\\
\> $\lambda$Q$\lambda$v(Q($\lambda$u$^\prime$[ $\mid$ save(v,u$^\prime$)]))
dr(ant(it$_3$)) \\
\end{tabbing}

We substitute the value of x$_3$ for $dr(ant(it_3))$:

\begin{tabbing}
xxx\= xxx\= \kill
save it$_3$  $\Rightarrow$\\

\> $\lambda$Q$\lambda$v(Q($\lambda$u$^\prime$[ $\mid$ save(v,u$^\prime$)])) \\
$\lambda$P([u$_3$ $\mid$ of(u$_3$,u$_0$),paycheck(u$_3$)];P(u$_3$))]
\\
\end{tabbing}

We perform $\lambda$ reductions, resulting in:
\begin{tabbing}
xxx\= xxx\= \kill
save it$_3$  $\Rightarrow$\\
\> $\lambda$v ([u$_3$ $\mid$ of(u$_3$,u$_0$),paycheck(u$_3$)];\\
\>[ $\mid$ save(v,u$_3$)])) \\

Jones$_4$* PAST$_5$ save it$_3$.  $\Rightarrow$\\

\> [u$_0$,u$_4$,P$_5$,u$_3$ $\mid$ u$_0$ = u$_4$,u$_4$=Jones,
 of(u$_3$,u$_0$),\\
\> \> paycheck(u$_3$), save(u$_4$,u$_3$)] \\
\end{tabbing}

The complete discourse is represented as follows:
\begin{tabbing}

\= xx P$_2$ [x$_3$ $\mid$ \= x$_3$ = \= \kill

Smith $_1$* PAST$_2$ spend his* paycheck$_3$.\\ 
Jones$_4$* PAST$_5$ save it$_3$.  $\Rightarrow$\\

\> [u$_0$,u$_1$,P$_2$,x$_3$ $\mid$ u$_0$ = u$_1$,u$_1$ = Smith,\\
\> \> x$_3$ = \\
\> \> $\lambda$P([u$_3$ $\mid$ of(u$_3$,u$_0$),paycheck(u$_3$)];P(u$_3$)) \\
\>   [u$_3$ $\mid$ of(u$_3$,u$_0$), paycheck(u$_3$),spend(u$_1$,u$_3$)]; \\
\> [u$_0$,u$_4$,P$_5$,u$_3$ $\mid$ u$_0$ = u$_4$,u$_4$=Jones,\\
\> \> of(u$_3$,u$_0$),paycheck(u$_3$), save(u$_4$,u$_3$)] \\
\end{tabbing}

The dynamic individual x$_3$ adds the paycheck of
u$_0$ (the discourse center) to the context.  In the second sentence, the
discourse center is $Jones$.  
Thus we get the reading in which ``Jones saved Jones' paycheck'', as desired.

\section{An Expanded Paradigm for Sloppy Identity}

The proposed theory permits a simple, uniform treatment of 
sloppy identity in VPE and paycheck pronouns.  This uniformity extends
further.  We simply permit sloppy identity
for any proform,
whenever the antecedent contains a proform within it.  This is
schematically represented as follows:

\bigskip

C1 \ldots [$_{XP}$ \ldots [$_{YP}$] \ldots] \ldots C2 \ldots
[$_{XP^{\prime}}$]

(C1, C2: ``controllers'' of sloppy variable {\bf YP})

\bigskip

Here, $XP$ is the antecedent for some proform $XP^\prime$, and $YP$ is the 
sloppy variable, i.e., a proform embedded within $XP$.  
A sloppy reading results whenever there is a {\em center shift}
involving C1 and C2.
That is, the interpretation of
$YP$ switches from controller $C1$ to $C2$.

Since the dynamic theory treats 
VP ellipsis uniformly with NP proforms, $XP$ and $YP$ both range over $NP$ and 
$VP$. This predicts four possibilities.
All four possibilities in fact occur, as shown by the following examples:
\beginsentences
\sitem Tom {\bf [$_{VP}$ loves [$_{NP}$ his] cat]}.  John does too. 
\label{standard}
\sitem Smith spent {\bf [$_{NP}$ [$_{NP}$ his] paycheck]}.
Jones saved it.
\label{lazy}
\sitem  I'll help you if you {\bf  [$_{VP}$ want me to  [$_{VP}$ ] ]}. 
I'll kiss you even if you don't.\footnote{This example was provided by
Marc
Gawron (p.c.), who attributed it to Carl Pollard.}
\label{poll}
\sitem  When Harry drinks, I always conceal {\bf [$_{NP}$ my belief that
he
shouldn't [$_{VP}$ ] ]}. 
When he gambles, I can't conceal it.
\label{last}
\endsentences

Examples \ref{standard} and \ref{lazy} have already been discussed.
\ref{standard} is the familiar case in which the VP antecedent ($XP$) 
contains a sloppy pronoun ($YP$). $YP$ switches from $C1$, {\em Tom}, to 
$C2$, {\em John}.  
In example \ref{lazy}, we have an NP antecedent ($XP$) containing a 
sloppy pronoun ($YP$), and
the two controllers for $YP$ are {\em Smith} and {\em Jones}. 
Example \ref{poll} 
involves a VP antecedent containing a sloppy VP ellipsis; the VP ellipsis 
switches from {\em help you} to {\em kiss you}.  Finally, example \ref{last} 
involves an NP antecedent containing a sloppy VP ellipsis, switching from 
{\em drinks} to {\em gambles}.  

We have already seen how the sloppy reading is derived for \ref{standard}
and for \ref{lazy}. We now show the derivation for \ref{poll}
(example \ref{last} can be derived in a similar fashion.)\footnote{We 
construct a representation as if the connectives
{\em if} and {\em even if} were simple conjunctions.  This allows us
to avoid the complex issues involved in representing such ``backwards
conditionals'' in a dynamic system.}:

\begin{tabbing}
 [u$_1$,P$_0$, \= P$_2$,u$_3$,P$_4$  $\mid$ \= u$_1$ = I,P$_0$ = P$_2$,u$_3$ = You,  \kill

I$_1$ WILL$_2$* help you$_3$ [if] you$_3$ PRES$_4$ want me$_1$ to$_2$. \\
I$_1$ WILL$_5$* kiss you$_3$ [even if] you$_3$ DO$_4$ NOT.
$\Rightarrow$ \\

 $[$u$_1$,P$_0$,P$_2$,u$_3$,P$_4$  $\mid$ u$_1$ = I,P$_0$ = P$_2$,u$_3$ = You, \\
\> \> P$_2$ = $\lambda$v([ $\mid$ help(v,u$_3$)$]$),\\
\> \>P$_4$ = $\lambda$v([ $\mid$ want(v,P$_0$(u$_1$))]), \\
 help(u$_1$,u$_3$),want(u$_1$,help(u$_1$,u$_3$))] ; \\

$[$P$_0$,P$_5$  $\mid$ P$_0$ = P$_5$, \\
\> P$_5$ = $\lambda$v([ $\mid$ kiss(v,u$_3$)]),NOT(P$_4$(u$_3$))$]$ \\

\end{tabbing}

The variable P$_4$ represents the property of ``wanting u$_1$ to P$_0$''.
Below, we substitute the value $\lambda$v([ $\mid$ want(v,P$_0$(u$_1$))]) for P$_4$, 
and then substitute the value $\lambda$v([ $\mid$ help(v,u$_3$)$]$) for
P$_0$, and apply it to u$_3$, giving the following result:

\begin{tabbing}
 [u$_1$,P$_0$,P$_2$,u$_3$,P$_4$  $\mid$ \= u$_1$ = I,P$_0$ = P$_2$,u$_3$ = You,  \kill

I$_1$ WILL$_2$* help you$_3$ [if] you$_3$ PRES$_4$ want me$_1$ to$_2$. \\
I$_1$ WILL$_5$* kiss you$_3$ [even if] you$_3$ DO$_4$ NOT.
$\Rightarrow$ \\
$[$u$_1$,P$_0$,P$_2$,u$_3$,P$_4$  $\mid$ u$_1$ = I,P$_0$ = P$_2$,u$_3$ = You,\\
\> P$_2$ = $\lambda$v([ $\mid$ help(v,u$_3$)$]$), \\
\> P$_4$ = $\lambda$v([ $\mid$ want(v,P$_0$(u$_1$))]), \\
\> help(u$_1$,u$_3$),want(u$_1$,help(u$_1$,u$_3$))] ; \\

$[$P$_0$,P$_5$  $\mid$ P$_0$ = P$_5$,
  P$_5$ = $\lambda$v([ $\mid$ kiss(v,u$_3$)]),\\
\>NOT([ $\mid$ want(u$_3$,kiss(u$_1$,u$_3$))]), \\

\end{tabbing}

It is the ``center shift'' involving P$_2$ (``help you'') and P$_5$ (``kiss you'')
that makes the desired reading possible.  That is, ``what u$_3$ doesn't want is for
u$_1$ to kiss u$_3$''.

The dynamic theory explains all
four of these cases in the same way; the embedded proform in the
antecedent can be sloppy, because
the controller for the embedded proform can undergo a {\em center shift}.
The cases illustrated by \ref{poll} and \ref{last} have not, to my knowledge,
been discussed previously in the literature.
It is not clear how such examples
could be handled by alternative theories, such as \cite{Fiengo&May:94} or
\cite{DSP:91}, since these theories do not treat NP and VP anaphora
in a uniform fashion.

\section{Conclusions and Future Work}

The dynamic perspective provides a framework for a simple, intuitive
account of sloppy identity and related phenomena, by explaining the
interpretive facts in terms of changes in context.  This requires 
contexts to change in a way that is somewhat foreign to the dynamic
perspective; a given position in the context must be reassigned,
or {\em shift} its value.  To implement this, I have incorporated
the notion of {\em discourse center}, together with the mechanism
of {\em center shift}, into a dynamic system.  This makes it
possible to give a novel, dynamic account of sloppy identity
phenomena.  I have shown that this approach accounts for
an expanded paradigm of sloppy identity, going beyond the 
data addressed in alternative accounts.  In future work, we
will investigate incorporating additional aspects of centering
theory, including the forward-looking centers list, and the
preference orderings on transitions.

\section{Acknowledgments}
Thanks to Claire Gardent, Aravind Joshi, Shalom Lappin, Mats Rooth,
Stuart Shieber, Mark Steedman, and Bonnie Webber for
help in developing the basic approach described in this paper.
Thanks to Reinhard Muskens for helpful comments on an earlier version of this work.
This work was partially supported by a Villanova University
Summer Research Grant (1995), and an NSF Career Grant, no. IRI-9502257.  

\bibliographystyle{acl}
\bibliography{/mnt/a/hardt/texfiles/bib.bib}

\end{document}